\documentclass[a4paper,conference]{IEEEtran}
\usepackage[table,xcdraw]{xcolor}
\usepackage{graphicx}
\usepackage{amsmath,amssymb,amsfonts}
\usepackage{amsfonts}
\usepackage{tikz}
\usepackage{enumitem}
\usepackage{array, makecell}
\usepackage{hhline}
\usepackage{algorithm}
\usepackage{algorithmic}
\usepackage{bm}
\usepackage{multirow}

\def\mycomment#1{}

\ifCLASSINFOpdf
 
\else
\fi
\begin{document}

\title{A Novel Adaptive Minority Oversampling Technique for Improved Classification in Data Imbalanced Scenarios}

\author{Ayush Tripathi, Rupayan Chakraborty, Sunil Kumar Kopparapu \\
 \IEEEauthorblockA{\textit{TCS Research, Tata Consultancy Services Ltd., INDIA}}
	\{t.ayush, rupayan.chakraborty, sunilkumar.kopparapu\}@tcs.com
}

\maketitle

\begin{abstract}

Imbalance in the proportion of training samples belonging to different classes
often poses performance degradation of conventional classifiers. This is
primarily due to the tendency of the classifier to be biased towards the
majority classes in the imbalanced dataset. In this paper, we propose a novel
three step technique to address imbalanced data. As a first step we
significantly oversample the minority class distribution by employing the
traditional Synthetic Minority Oversampling Technique (SMOTE) algorithm using
the neighborhood of the minority class samples and in the next step we
partition the generated samples using a Gaussian-Mixture Model based clustering
algorithm. In the final step synthetic data samples are chosen based on the
weight associated with the cluster, the weight itself being determined by the
distribution of the majority class samples. Extensive experiments on several
standard datasets from diverse domains show the usefulness of the proposed
technique in comparison with the original SMOTE and its state-of-the-art
variants algorithms.
\mycomment{
Class imbalance in any recognition task is a crucial problem where proportion of samples belonging to different classes differs in the training dataset.  As a result, the performance of conventional classifiers tends to be biased towards the majority classes in the imbalanced dataset. To deal with this problem, we propose a novel oversampling technique in this paper. In the first step, we significantly oversample the minority class distribution by employing the traditional Synthetic Minority Oversampling Technique (SMOTE) algorithm using the neighborhood of the minority class samples. Then in the second step, the generated samples are partitioned using a Gaussian-Mixture Model based clustering algorithm. And in the last step, the distribution of the majority class samples are utilized to assign a weight to each of the clusters from which the synthetic data samples are chosen based on the weight associated with the cluster. Extensive experiments on several standard datasets from diversified domains clearly indicate the usefulness of the proposed technique in comparison with the original SMOTE algorithm and its state-of-the-art variants.}
\end{abstract}


\IEEEpeerreviewmaketitle

\section{Introduction}

Class imbalance in statistical Machine Learning (ML) is a problem in which number of samples corresponds to each class is not proportionate. It means that one
or more class, called the minority class, is under represented with few number 
of samples compared to other class with comparatively higher number of samples, called the majority class. During the past two decades, the class imbalance 
problem has been observed in a lot of real-world applications, including 
fraud detection \cite{fraud}, network intrusion detection \cite{8526292}, 
disease diagnosis \cite{8246503}, software defect detection \cite{10.1145}, 
industrial manufacturing \cite{KIRLIDOG2012989}, bioinformatics, environment resource management, security management to
name a few. 
The performance of conventional classifiers such as Support Vector Machines 
(SVMs), Artificial Neural Networks (ANNs) etc., in class imbalance scenario, tends to be biased towards the 
majority class.

Various approaches have been proposed in literature to deal with classification in class imbalanced scenarios \cite{10.1145}. Sampling \cite{SMOTE,4717268} is the most commonly used approach, which involves readjusting the distribution of the training set to achieve a balanced class condition. In contrast to other 
techniques, sampling approach has the advantage of being independent of the underlying classification model.  Ensemble learning is yet another popular approach \cite{6609011} adopted to tackle the class imbalance problem;
by fusing multiple classifiers, either by using voting or combining the classifier scores. However, choosing an optimal combination of ensemble of classifiers is still an open problem. Cost-sensitive learning based methods \cite{Krawczyk2014CostsensitiveDT} use an appropriate cost parameter for minority and majority class samples. But selecting a proper value of the cost parameter is critical for the performance of these methods. Additionally, data representative methods like coordination learning 
and simultaneous two sample (s2s) learning \cite{ijcai2018-290} have also been used to tackle the class imbalance problem.

Synthetic Minority Oversampling Technique (SMOTE) is one of the most popular and well-known sampling algorithms for addressing class imbalance in
machine learning. However, it is observed that local distribution characteristics of the data are often ignored during the oversampling process by SMOTE. 
In this paper, we propose a novel minority oversampling technique which makes
use of the local distribution of the data, to address the class imbalance 
problem. For this purpose, we follow a two step oversampling followed by an undersampling step. In the oversampling step, we use SMOTE to generate a
 pre-specified number of synthetic samples. The number of samples to be 
generated is empirically chosen such that it is much larger than that 
required for a balanced distribution. This is followed by an undersampling 
procedure to adaptively select a set of samples that not only (a) mitigate the 
woes of class imbalance, but also (b) ensures that the minority samples are 
diverse enough to maintain the original data distribution. 
For this, we first cluster the samples generated by SMOTE, by using a 
Gaussian-Mixture Model (GMM) based clustering algorithm. 
Each of the clusters is then weighted by utilizing the distribution of the majority class samples. Samples are then chosen from each of the clusters based on the weights assigned to it. Though hybrid sampling techniques, involving a combination of oversampling and undersampling methods, have been proposed in 
literature, to the best of our knowledge, the use of probability density 
estimation strategy based hybrid sampling has not been explored. This is the
main contribution of this paper. We tested the proposed technique on a large
($27$)
number of binary imbalanced datasets, from different domains with varied 
class imbalance ratios.
The results indicate that the proposed algorithm outperforms SMOTE and other state-of-the-art oversampling techniques.

The rest of the paper is organized as follows. In Section \ref{sec:prior}, we review the existing literature on techniques used to deal with class imbalance scenarios. We motivate and describe the proposed setup in Section \ref{sec:proposed}.  In Section \ref{sec:expt} we
analyze  the experimental outcomes and conclude in Section \ref{sec:conclusion}. 

\section{Prior Work} \label{sec:prior}

Several strategies have been proposed in literature to tackle the class
imbalance problem. Broadly, these methods fall into one of the 
three categories \cite{HAIXIANG2017220}, namely, (a) Resampling techniques, (b) Cost sensitive learning and (c) Ensemble based classifiers.


\noindent \textbf{Resampling Techniques}: This family of techniques, aims at re-balancing the sample space, given an imbalanced dataset with the aim of alleviating the effects of skewness in class distribution in the learning process. Such methods are versatile as they are independent of the selection of a suitable classifier. These techniques can be further classified into three groups based on the approach used to achieve a balanced distribution of samples,
namely,
	
\begin{enumerate}

	\item OverSampling: The process of achieving a balanced data distribution by increasing the minority class instances is referred to as oversampling. Random OverSampling (ROS), which is the process of randomly duplicating 
the minority class samples is widely used to alleviate class imbalance. The Synthetic Minority Oversampling Technique (SMOTE) \cite{SMOTE}, inserts a synthetic minority class instance between two randomly set
of adjacent minority class examples. In \cite{BSMOTE}, the authors proposed the Borderline-SMOTE (B-SMOTE) algorithm which first divides the minority class instances into safety, danger and noisy groups and subsequently apply the SMOTE algorithm only in the danger group. Subsequently, they proposed two versions of B-SMOTE,
namely, B-SMOTE1 and B-SMOTE2. In B-SMOTE1, new instances are generated between the borderline minority class samples and their nearest neighbors belonging to the same class, while in B-SMOTE2, the majority class instances are also 
used for the neighborhood calculation. Following a similar path, various other variations of the SMOTE algorithm has been proposed in recent literature,
namely,  Safe-Level-SMOTE \cite{SLSMOTE}, SMOTE-IPF \cite{SMOTEIPF}, Surrounding Neighborhood-based SMOTE (SN-SMOTE) \cite{SNSMOTE}, Grouped-SMOTE \cite{GSMOTE}, Majority  Weighted  Minority  Oversampling  Technique (MW-MOTE) \cite{MWMOTE}.

	\item UnderSampling: This process involves eliminating the class imbalance problem by discarding samples of the majority class. A simple, yet effective method to achieve this is Random Under Sampling (RUS) \cite{RUS}, which aims at achieving a balanced distribution by randomly eliminating the majority class examples. The NearMiss technique \cite{NearMiss}, performs undersampling of points in the majority class based on the distance of samples to other samples belonging to the same class. Three different versions of NearMiss have been proposed. In the NearMiss-1, the examples belonging to the majority class that have the smallest average distance to the three closest minority class samples are retained. NearMiss-2 retains the majority class samples that have the smallest average distance to three furthest minority class samples
and NearMiss-3 involves selecting a specified number of majority class examples closest to each minority class sample. Condensed Nearest Neighbors (CNN) \cite{CNN} is an undersampling technique that aims to find a suitable subset (referred to as the minimal consistent set) of the majority class samples that result in optimal classifier performance. On the same grounds, various other undersampling techniques such as Edited Nearest Neighbor (ENN), Tomek Link Removal, Neighborhood Cleaning Rule have been proposed in the literature \cite{5211614}.

	\item Hybrid Sampling: These set of techniques involve using a combination of oversampling and undersampling techniques in order to obtain a balanced distribution. The most widely used hybrid techniques are SMOTE followed by Tomek Link Removal and SMOTE with Edited Nearest Neighbor (ENN) based undersampling \cite{4583030}.
		
\end{enumerate}

\noindent \textbf{Cost Sensitive Learning}: Cost-sensitive learning based methods aim at dealing with class imbalance problem by assigning higher misclassification cost to the minority class samples compared to the majority class samples \cite{6469237,Lopez2015CostsensitiveLF}. In spite of being more computationally efficient as compared to sampling based methods, cost sensitive learning is less popular due to the difficulty involved in precise assignment of misclassification costs \cite{Krawczyk2014CostsensitiveDT}. There are three main approaches for dealing with cost-sensitive problems. The first set of methods are based on training data modification, where the decision thresholds are modified based on the training data distribution \cite{7567497,7552905}. The second set of approaches involves changing the learning process to build a cost-sensitive classifier. This can be achieved by modifying the objective function of the classifier by using a weighting strategy or by defining an appropriate error function on the neural network \cite{4695979,8545796}. Lastly, incorporating the cost matrix into a Bayesian decision boundary has also been explored in recent literature \cite{Datta,6784638}.

\noindent \textbf{Ensemble Based Classifiers} : Ensemble-based classifiers aim to improve the classification performance of a single classifier by using a combination of multiple base classifiers \cite{6609011}. Ensemble models can be categorized into two categories - iterative based ensemble and parallel based ensemble. Parallel based ensemble refers to the process of training each base classifier in parallel and includes bagging, re-sampling based ensemble and feature selection based ensemble techniques \cite{HAO2014117,CAO2014137}. As suggested in \cite{5978225}, re-sampling based methods are often combined with parallel based ensemble schemes. Iterative based ensemble schemes employ bagging and boosting techniques to enhance classifier performance. In the Ada-boost algorithm \cite{10.5555}, the correct class samples are given higher weights, thereby forcing the classifier to focus more on learning the failed classified samples. Several other boosting techniques such as EHSBoost \cite{EHSBoost}, RUSBoost \cite{RUSBoost}, AdaBoost.M1, AdaBoost.M2 have been found to be efficient in dealing with imbalanced classification tasks.

\section{Our Approach} \label{sec:proposed}

As mentioned earlier, SMOTE is the most  popular and widely used technique for oversampling the minority class. The underlying idea is to interpolate and generate artificial samples 
so that they lie in close proximity of a pair of minority samples. 
Despite the advantage of SMOTE in alleviating the problem of over-fitting when compared to random oversampling, it suffers from some significant drawbacks. The first one is that owing to the randomness involved in the algorithm, it leads to some interpolated samples appearing at 
inappropriate locations in the sample space. Also, it has been observed that noisy information in the original dataset is propagated further in the oversampling procedure leading to deterioration in classifier performance. Thus, motivated by these drawbacks of the conventional SMOTE algorithm, we propose a $3-step$ procedure for addressing the aforementioned shortcomings. 
The proposed approach is shown in Algorithm \ref{algorithm1:minority_upsampling} and can
be described in three steps.

\begin{algorithm}
	\caption{Adaptive minority up-sampling}
	\label{algorithm1:minority_upsampling}
	\textbf{Input}:  $M (N) \rightarrow$ \# of majority (minority) class samples  \\ 
	\textbf{Output}: $N_u \rightarrow$ \# of Upsampled minority class samples   \\
	\textbf{Step 1: Sample Generation}\\
	$d$ = distance operator, $S_{p}$ = sample pick-up operator, $J$ = sample joining operator, $r$ = \# of samples to be picked up, $K$ = \# of nearest neighbors
	\begin{algorithmic}[1] 
		\STATE $\hat{N}_1=0$ \\
		\FOR{$n$ in \{1, 2, ..., $N$\}}
		\STATE $\chi \leftarrow (d(s_{n}, s_{l}),K)$, 
		where $\forall\ l \neq n$, $1 \leq l \leq K$, \\ 
		where $\chi$ contains $K$ neighbors of $s_n$ (Ball-Tree algorithm)
		\FOR{$k$ in \{1, 2, ..., $K$\}}
		\STATE $\tilde{N} \leftarrow S_{p}\left (J(s_n,s_k),r \right)$
		\STATE $\hat{N}_1 = \hat{N}_1 + \tilde{N}$S
		\ENDFOR
		\ENDFOR \\
	\end{algorithmic}
	\textbf{Step 2: GMM Clustering} \\
	\textbf{Input}: $N+\hat{N}_1$ = Total minority samples for clustering,\\ $C$ = Number of clusters 
	\begin{algorithmic}[1]
		\STATE $GMM_{C} \left ((N+\hat{N}_1), C\right)$
	\end{algorithmic}
	\textbf{Step 3: Adaptive Sample Selection} \\
	$S_{C}^{i}$: Selection operator picking up samples from $\tilde{N}$ points belonging to the same cluster $C$, \\ $ p(m_{j} \in C_{i} | M)$ = posterior probability of a majority class sample $m_{j}$ belonging to the cluster $C_{i}$\\ $p_{t}$ = probability threshold, $W$ = weight array, \\ $w_{t}$ = weight threshold
	\begin{algorithmic}[1]
		\STATE $W=[0]$
		\FOR {$i$ in \{1, 2, ..., $C$\}}
		\STATE $q=0$
		\FOR {$j$ in \{1, 2, ..., $M$\}}
		\IF {$ p(m_{j} \in M | C_{i}) > p_{t}$}
		\STATE $q=q+1$
		\ENDIF
		\ENDFOR
		\IF {$(w_{i}=(1- \frac{q}{M}))> w_{t}$}
		\STATE $W = [ W, w_{i} ]$
		\ELSE
		\STATE $W= [W,0]$
		\ENDIF
		\ENDFOR
		\FOR {$i$ in \{1, 2, ..., $C$\}}
		\STATE $\hat{N}_2 \leftarrow S_{C}^{i} \Bigg(\Big((M-N)\times\frac{W_{i}}{\sum_{i=1}^{C}W_{i}}\Big),N\Bigg) $
		\STATE $N_u \leftarrow \hat{N}_2 + N $
		\ENDFOR
	\end{algorithmic}
\end{algorithm}

\noindent \textbf{Step 1: Sample Generation}: This is the preliminary step and identical to the SMOTE oversampling technique. In this step, for a given minority class sample $s_n$, 
we find its $K$ nearest neighbors within the minority class ($s_1, s_2, \cdots,
s_N$) using the Ball Tree algorithm. 
Next, we generate a new set of samples between the sample $s_n$ and its neighbor $s_k$
for $k= 1, 2, \cdots, K$ by using the distance operator ($d$) as, 
$s'_n = s_n + \alpha \times (s_n-s_k)$ 
where  $\alpha$ takes $r$ values, namely, $\frac{1}{r}, \frac{2}{r}, \cdots,
\frac{r}{r}$  obtained by dividing the interval $(0,1)$ into $r$ equal divisions. 
This essentially leads to the generation of $r$ synthetic samples for each 
"minority sample ($s_n$) and neighbor ($s_k$)" pair. 
By repeating this process for each of the $N$ minority samples (line $4-7$), we obtain an intermediate oversampled distribution consisting of $\hat{N}_1$ additional minority samples. 

\noindent \textbf{Step 2: GMM Clustering}: The minority distribution consisting of the $N$ original samples and $\hat{N}_1$ synthetic samples is now subjected to a Gaussian-Mixture Model based clustering algorithm. In order to estimate the GMM parameters, we use the Expectation-Maximization algorithm. As a result of this step, the ($N + \hat{N}_1$) samples are clustered into $C$ clusters, wherein each cluster is modeled according to a Gaussian distribution. 

\noindent \textbf{Step 3: Adaptive Sample Selection}: In this step, we utilize the majority sample distribution to adaptively select the required set of samples ($N_u$) to achieve a balanced distribution from the generated minority samples ($\hat{N}_1$). For each of the 
$m = 1, 2, \cdots M$ majority class samples, we compute the posterior probability of the 
sample belonging to the $i^{th}$ cluster $C_i$ given as $p(m_{j} \in C_{i} | M)$. 
In order to assign a weight to the $i^{th}$ cluster, we adopt the following strategy:
\begin{itemize}
	\item We define a probability threshold parameter ($p_t$) and compute the number of majority class samples for which the posterior probability, $p(m_{j} \in C_{i} | M)$ exceeds $p_t$. Suppose the number of such points for the cluster is $q_i$. 
	\item Now, each cluster is assigned a weight $w_i$ as, 
	\[
	w_i= 
	\begin{cases}
	(1-\frac{q_i}{M}),& \text{if } w_i>w_t\\
	0,              & \text{otherwise}
	\end{cases}
	\]
	where, $w_t$ is a weight threshold parameter.
\end{itemize}

Next, we identify the number of samples to be chosen from each cluster by,
\[N_i = \Big((M-N)\times\frac{w_{i}}{\sum_{i=1}^{C}w_{i}}\Big)
\] 
Finally, we select $N_i$ points belonging to the synthetically generated samples $\hat{N}_1$ lying in the $i^{th}$ cluster. 

Thus, by using the technique described in Algorithm \ref{algorithm1:minority_upsampling}, we generate $(M-N)$ additional minority class samples, which leads to a balanced distribution. The use of GMM-based clustering and weighting strategy ensures that the local data distribution characteristics are also captured. Moreover, by having a proper choice of the probability and weight threshold parameters, it is ensured that the noise information (if present) in the original data distribution does not get propagated during the oversampling process.

\section{Experiments and Results} 
\label{sec:expt}

\subsection{Dataset Description}

In order to validate the efficacy of the proposed algorithm, we conducted experiments on a 
wide variety of binary-class imbalanced datasets, belonging to different domains. 
This includes $10$ standard datasets taken from the Keel dataset repository \cite{Keel}, 
noisy and borderline examples \cite{Noisy}, speech emotion recognition \cite{10.5555/3207761} 
and audio dementia classification \cite{8903138,our_EUSIPCO}.

The detailed information about the data sets obtained from the Keel dataset repository is provided in Table \ref{table:keel}. These datasets include Pima, Ecoli, Yeast and Abalone (computational biology), Pageblock (document classification), Vehicle (image annotation) and Glass (object classification). For noisy and borderline examples, we consider the 04clover5z dataset \cite{Noisy} which is an artificial dataset consisting of $600$ ($500$ majority and $100$ minority samples, with imbalance ratio of $5$) examples randomly and uniformly distributed in a two-dimensional space with real-valued attributes. The minority class resembles a flower with elliptic petals and the borders of sub-regions in the minority class are disturbed. We perform experiments on five such datasets with the disturbance ratio varying from $0$ to $70$. 

We consider the standard Berlin speech emotion database (EMO-DB) \cite{Burkhardt2005ADO} consisting of $535$ utterances corresponding to $7$ different emotions (Anger - $127$, Boredom - $81$, Neutral - $79$, Happy - $71$, Anxiety - $69$, Sad - $62$ and Disgust - $46$) for the task of emotion classification. Each utterance is represented by a $62$-dimensional feature vector obtained by using the GeMAPSv01a configuration using the openSMILE toolkit \cite{opensmile}. For two class classification, we consider $10$ different imbalanced scenarios from the EMO-DB as described in Table \ref{table:emodb}. For the task of dementia classification we use the Pitt Corpus \cite{Pitt} which comprises of $255$ AD (Alzheimer's Disease), $242$ HC (Healthy Control) and $116$ MCI (Mild Cognitive Impairment) subjects. We consider the AD-MCI and HC-MCI tasks for validating the proposed algorithm with an imbalance ratio of $2.19$ and $2.08$ respectively. The feature set used for classification is same as that used for the emotion classification task. Please note that our main intention of the experiments for emotion and dementia classification is not to identify the most appropriate set of features but to demonstrate the effectiveness of our approach in class imbalanced scenarios.

To summarize, we consider a total of $27$ different datasets encompassing various domains and having different number of attributes, number of instances and class imbalance ratios.

\begin{table}
	\caption{Details of the standard imbalanced datasets taken from the Keel dataset repository.}
	\centering
	\scalebox{0.9}{
	\begin{tabular}{|l|c|c|c|c|}
		\hline
		\textbf{Dataset} & \textbf{Attributes} & \textbf{Majority} & \textbf{Minority} & \textbf{Imbalance Ratio} \\ \hline \hline
		Pima                                    & 8                                          & 500               & 268               & 1.90                     \\ \hline
		Glass0                                  & 9                                          & 144               & 70                & 2.01                     \\ \hline
		Vehicle0                                & 18                                         & 647               & 199               & 3.23                     \\ \hline
		Ecoli1                                  & 7                                          & 259               & 77                & 3.36                     \\ \hline
		Yeast3                                  & 8                                          & 1321              & 163               & 8.11                     \\ \hline
		Pageblock                               & 10                                         & 444               & 28                & 15.85                    \\ \hline
		Glass5                                  & 9                                          & 205               & 9                 & 22.81                    \\ \hline
		Yeast5                                  & 8                                          & 1440              & 44                & 32.78                    \\ \hline
		Yeast6                                  & 8                                          & 1449              & 35                & 39.15                    \\ \hline
		Abalone                                 & 8                                          & 4142              & 32                & 128.67                   \\ \hline
	\end{tabular}
}
\label{table:keel}
\end{table} 

\begin{table}
	\caption{Details of the datasets for imbalanced emotion classification task.}
	\centering
	\scalebox{1.0}{
		\begin{tabular}{|l|c|c|c|}
			\hline
			\textbf{Classification Task} &  \textbf{Majority} & \textbf{Minority} & \textbf{Imbalance Ratio} \\ \hline \hline
			Anger-Anxiety                                       & 127                                      & 69                & 1.84                     \\ \hline
			Anger-Disgust                                       & 127                                      & 46                & 2.76                     \\ \hline
			Anger-Happy                                         & 127                                      & 71                & 1.78                     \\ \hline
			Anger-Sad                                           & 127                                      & 62                & 2.05                     \\ \hline
			Anxiety-Disgust                                     & 69                                       & 46                & 1.5                      \\ \hline
			Boredom-Disgust                                     & 81                                       & 46                & 1.76                     \\ \hline
			Happy-Disgust                                       & 71                                       & 46                & 1.54                     \\ \hline
			Neutral-Anger                                       & 127                                      & 79                & 1.61                     \\ \hline
			Neutral-Disgust                                     & 79                                       & 46                & 1.72                     \\ \hline
			Sad-Disgust                                         & 62                                       & 46                & 1.35                     \\ \hline
		\end{tabular}
	}
	\label{table:emodb}
\end{table}

\begin{table*}[]
	\caption{$F_1$ scores of the comparative algorithms.}
	\centering
	\scalebox{0.77}{
		\begin{tabular}{|l|c|c|c|c|c|c|c|c|}
			\hline
			\textbf{Classification Task} & \textbf{Baseline} & \textbf{ROS}   & \textbf{SMOTE}         & \textbf{BSMOTE1}       & \textbf{BSMOTE2}       & \textbf{SVMSMOTE}      & \textbf{ADASYN}        & \textbf{Proposed}      \\ \hline \hline
			\textbf{Pima}                                       & 0.6167 $\pm$ 0.0024                            & 0.6720 $\pm$  $\pm$ 0.0009 & 0.6794 $\pm$ 0.0023          & \textbf{0.6828 $\pm$ 0.0010} & 0.6769 $\pm$ 0.0023          & 0.6777 $\pm$ 0.0016          & \textbf{0.6834 $\pm$ 0.0015} & 0.6727 $\pm$ 0.0023          \\ \hline
			\textbf{Glass0}                                     & 0.6478 $\pm$ 0.0117                            & 0.7223 $\pm$ 0.0050  & 0.7449 $\pm$ 0.0061          & 0.7255 $\pm$ 0.0047          & 0.7246 $\pm$ 0.0031          & 0.7424 $\pm$ 0.0018          & 0.7153 $\pm$ 0.0047          & \textbf{0.7530 $\pm$ 0.0025}  \\ \hline
			\textbf{Vehicle0}                                   & 0.9324 $\pm$ 0.0005                            & 0.9234 $\pm$ 0.0002  & 0.9386 $\pm$ 0.0001          & 0.9227 $\pm$ 0.0002          & 0.9055 $\pm$ 0.0003          & 0.9259 $\pm$ 0.0003          & 0.9230 $\pm$ 0.0002          & \textbf{0.9498 $\pm$ 0.0004} \\ \hline
			\textbf{Ecoli1}                                     & 0.8171 $\pm$ 0.0046                            & 0.7680 $\pm$ 0.0036  & 0.7941 $\pm$ 0.0046          & \textbf{0.7812 $\pm$ 0.0051} & 0.7669 $\pm$ 0.0026          & 0.7765 $\pm$ 0.0034          & 0.7684 $\pm$ 0.0035          & \textbf{0.8261 $\pm$ 0.0046} \\ \hline
			\textbf{Yeast3}                                     & \textbf{0.7591 $\pm$ 0.0023}                   & 0.7162 $\pm$ 0.0029  & 0.7292 $\pm$ 0.0021          & 0.7231 $\pm$ 0.0019          & 0.6587 $\pm$ 0.0022          & \textbf{0.7518 $\pm$ 0.0010} & 0.6968 $\pm$ 0.0016          & \textbf{0.7591 $\pm$ 0.0020} \\ \hline
			\textbf{Pageblock}                                  & 0.5615 $\pm$ 0.0591                            & 0.876 $\pm$ 0.0058   & 0.9006 $\pm$ 0.0045          & \textbf{0.9446 $\pm$ 0.0061} & 0.9177 $\pm$ 0.0028          & 0.9006 $\pm$ 0.0045          & 0.9356 $\pm$ 0.0010          & \textbf{0.9600 $\pm$ 0.0063}   \\ \hline
			\textbf{Glass5}                                     & 0.6333 $\pm$ 0.1377                            & 0.6333 $\pm$ 0.1377  & 0.6333 $\pm$ 0.1377          & 0.6333 $\pm$ 0.1377          & 0.6333 $\pm$ 0.1377          & 0.6133 $\pm$ 0.1447          & 0.6133 $\pm$ 0.1447          & \textbf{0.6666 $\pm$ 0.1333} \\ \hline
			\textbf{Yeast5}                                     & 0.5762 $\pm$ 0.0155                            & 0.6284 $\pm$ 0.0017  & 0.6345 $\pm$ 0.0040          & 0.6493 $\pm$ 0.0014          & 0.4454 $\pm$ 0.0029          & 0.6337 $\pm$ 0.0029          & 0.6547 $\pm$ 0.0021          & \textbf{0.6853 $\pm$ 0.0040} \\ \hline
			\textbf{Yeast6}                                     & 0.2898 $\pm$ 0.0566                            & 0.3947 $\pm$ 0.0037  & 0.4171 $\pm$ 0.0058          & 0.4989 $\pm$ 0.0055          & 0.3850 $\pm$ 0.0081          & 0.4980 $\pm$ 0.0072          & 0.3393 $\pm$ 0.0049          & \textbf{0.5219 $\pm$ 0.0107} \\ \hline
			\textbf{Abalone}                                    & 0.0 $\pm$ 0.0                                  & 0.0547 $\pm$ 0.0002  & 0.0571 $\pm$  $\pm$ 0.0003         & 0.0415 $\pm$ 0.0004          & 0.0541 $\pm$ 0.0008          & 0.0544 $\pm$  $\pm$ 0.0007         & 0.0577 $\pm$ 0.0003          & \textbf{0.0688 $\pm$ 0.0002} \\ \hline
			\textbf{Anger-Anxiety}                              & 0.9157 $\pm$ 0.0009                            & 0.9155 $\pm$ 0.0002  & 0.9426 $\pm$ 0.0002          & 0.9426 $\pm$ 0.0002          & 0.9150 $\pm$ 0.0007          & \textbf{0.9384 $\pm$ 0.0012} & 0.9362 $\pm$ 0.0005          & \textbf{0.9501 $\pm$ 0.0002} \\ \hline
			\textbf{Anger-Disgust}                              & 0.8819 $\pm$ 0.0015                            & 0.8951 $\pm$ 0.0020  & 0.8858 $\pm$ 0.0025          & 0.8858 $\pm$ 0.0025          & 0.8886 $\pm$ 0.0025          & \textbf{0.9016 $\pm$ 0.0064} & 0.8755 $\pm$ 0.0031          & 0.8952 $\pm$ 0.0021          \\ \hline
			\textbf{Anger-Happy}                                & 0.6787 $\pm$ 0.0055                            & 0.7111 $\pm$ 0.0030  & 0.7051 $\pm$ 0.0022          & 0.6827 $\pm$ 0.0007          & 0.6727 $\pm$ 0.0037          & 0.7198 $\pm$ 0.0011          & 0.6976 $\pm$ 0.0013          & \textbf{0.7338 $\pm$ 0.0027} \\ \hline
			\textbf{Anger-Sad}                                  & 0.9770 $\pm$ 0.0008                            & 0.9818 $\pm$ 0.0013  & \textbf{0.9913 $\pm$ 0.0003} & \textbf{0.9913 $\pm$ 0.0003} & 0.9738 $\pm$ 0.0012          & \textbf{0.9777 $\pm$ 0.0008} & \textbf{0.9913 $\pm$ 0.0003} & \textbf{0.9913 $\pm$ 0.0003} \\ \hline
			\textbf{Anxiety-Disgust}                            & 0.8014 $\pm$ 0.0213                            & 0.8194 $\pm$ 0.0075  & 0.8482 $\pm$ 0.0086          & 0.8482 $\pm$ 0.0086          & 0.8482 $\pm$ 0.0086          & \textbf{0.8458 $\pm$ 0.0136} & 0.8482 $\pm$ 0.0086          & \textbf{0.8570 $\pm$ 0.0064} \\ \hline
			\textbf{Boredom-Disgust}                            & 0.8695 $\pm$ 0.0013                            & 0.8498 $\pm$ 0.0007  & 0.8845 $\pm$ 0.0001          & \textbf{0.8978 $\pm$ 0.0006} & \textbf{0.8926 $\pm$ 0.0021} & 0.8573 $\pm$ 0.0005          & 0.8875 $\pm$ 0.0015          & \textbf{0.8978 $\pm$ 0.0006} \\ \hline
			\textbf{Happy-Disgust}                              & 0.8974 $\pm$ 0.0021                            & 0.8955 $\pm$ 0.0040  & \textbf{0.9210 $\pm$ 0.0020} & 0.9079 $\pm$ 0.0035          & 0.9222 $\pm$ 0.0021          & \textbf{0.9261 $\pm$ 0.0027} & 0.9177 $\pm$ 0.0044          & \textbf{0.9210 $\pm$ 0.0020}  \\ \hline
			\textbf{Neutral-Anger}                              & 0.9353 $\pm$ 0.0021                            & 0.9280 $\pm$ 0.0023  & 0.9418 $\pm$ 0.0027          & 0.9289 $\pm$ 0.0014          & 0.9328 $\pm$ 0.0058          & \textbf{0.9426 $\pm$ 0.0013} & 0.9353 $\pm$ 0.0021          & 0.9353 $\pm$ 0.0021          \\ \hline
			\textbf{Neutral-Disgust}                            & 0.8472 $\pm$ 0.0033                            & 0.8517 $\pm$ 0.0016  & \textbf{0.8763 $\pm$ 0.0014} & 0.8613 $\pm$ 0.0024          & \textbf{0.8763 $\pm$ 0.0014} & 0.8660 $\pm$ 0.0019          & \textbf{0.8763 $\pm$ 0.0014} & 0.8646 $\pm$ 0.0003          \\ \hline
			\textbf{Sad-Disgust}                                & 0.9514 $\pm$ 0.0021                            & 0.9514 $\pm$ 0.0021  & 0.9514 $\pm$ 0.0021          & 0.9514 $\pm$ 0.0021          & 0.9514 $\pm$ 0.0021          & 0.9514 $\pm$ 0.0021          & 0.9514 $\pm$ 0.0021          & \textbf{0.9632 $\pm$ 0.0024} \\ \hline
			\textbf{AD-MCI}                                     & 0.5509 $\pm$ 0.0110                            & 0.5521 $\pm$ 0.0063  & 0.5507 $\pm$ 0.0037          & 0.5484 $\pm$ 0.0047          & 0.5533 $\pm$ 0.0060          & 0.5654 $\pm$ 0.0068          & 0.5609 $\pm$ 0.0067          & \textbf{0.5773 $\pm$ 0.0081} \\ \hline
			\textbf{HC-MCI}                                     & 0.5955 $\pm$ 0.0025                            & 0.5953 $\pm$ 0.0014   & 0.5900 $\pm$ 0.0025             & 0.5894 $\pm$ 0.0006           & 0.5919 $\pm$ 0.0024           & 0.6013 $\pm$ 0.0008           & 0.5937 $\pm$ 0.0023           & \textbf{0.6220 $\pm$ 0.0033}   \\ \hline
			\textbf{04clover5z-600-5-70-BI}                     & 0.0 $\pm$ 0.0                                  & 0.5007 $\pm$ 0.0012  & 0.5119 $\pm$ 0.0004          & 0.5164 $\pm$ 0.0017          & 0.5148 $\pm$ 0.0012          & 0.4983 $\pm$ 0.0073          & \textbf{0.5207 $\pm$ 0.0011} & \textbf{0.5268 $\pm$ 0.0041} \\ \hline
			\textbf{04clover5z-600-5-60-BI}                     & 0.0 $\pm$ 0.0                                  & 0.5133 $\pm$ 0.0044  & 0.5322 $\pm$ 0.0011          & 0.5188 $\pm$ 0.0013          & 0.5377 $\pm$ 0.0035          & 0.5322 $\pm$ 0.0041          & \textbf{0.5325 $\pm$ 0.0024} & \textbf{0.5414 $\pm$ 0.0017} \\ \hline
			\textbf{04clover5z-600-5-50-BI}                     & 0.0 $\pm$ 0.0                                  & 0.5234 $\pm$ 0.0038  & 0.5548 $\pm$ 0.0023          & 0.5372 $\pm$ 0.0015          & 0.5383 $\pm$ 0.0019          & 0.5166 $\pm$ 0.0033          & 0.5426 $\pm$ 0.0012          & \textbf{0.5493 $\pm$ 0.0024} \\ \hline
			\textbf{04clover5z-600-5-30-BI}                     & 0.0 $\pm$ 0.0                                  & 0.5117 $\pm$ 0.0018  & 0.5227 $\pm$ 0.0020          & 0.5172 $\pm$ 0.0011          & 0.5226 $\pm$ 0.0004          & 0.5248 $\pm$ 0.0018          & 0.5288 $\pm$ 0.0022          & \textbf{0.5419 $\pm$ 0.0003} \\ \hline
			\textbf{04clover5z-600-5-0-BI}                      & 0.0 $\pm$ 0.0                                  & 0.5210 $\pm$ 0.0004  & 0.5311 $\pm$ 0.0011          & 0.5533 $\pm$ 0.0022          & 0.5635 $\pm$ 0.0015          & 0.5434 $\pm$ 0.0009          & 0.5533 $\pm$ 0.0009          & \textbf{0.5674 $\pm$ 0.0004} \\ \hline
		\end{tabular}
	}
	\label{table:F1}
\end{table*}

\begin{table*}
	\caption{$F_2$ scores of the comparative algorithms.}
	\centering
	\scalebox{0.8}{
		\begin{tabular}{|l|c|c|c|c|c|c|c|c|}
			\hline
			\textbf{Classification Task} & \textbf{Baseline} & \textbf{ROS}  & \textbf{SMOTE}         & \textbf{BSMOTE1}       & \textbf{BSMOTE2}       & \textbf{SVMSMOTE}      & \textbf{ADASYN}        & \textbf{Proposed}      \\ \hline \hline
			\textbf{Pima}                                       & 0.5684 $\pm$ 0.0029                            & 0.7039 $\pm$ 0.0013 & 0.7155 $\pm$ 0.0031          & \textbf{0.7414 $\pm$ 0.0021} & 0.7346 $\pm$ 0.0030          & 0.7143 $\pm$ 0.0032          & \textbf{0.7378 $\pm$ 0.0022} & 0.6892 $\pm$ 0.0032          \\ \hline
			\textbf{Glass0}                                     & 0.6262 $\pm$ 0.0185                            & 0.8047 $\pm$ 0.0058 & 0.8158 $\pm$ 0.0064          & 0.8206 $\pm$ 0.0048          & 0.8189 $\pm$ 0.0049          & 0.8262 $\pm$ 0.0012          & 0.8143 $\pm$ 0.0042          & \textbf{0.8267 $\pm$ 0.0024} \\ \hline
			\textbf{Vehicle0}                                   & 0.9367 $\pm$ 0.0011                            & 0.9649 $\pm$ 0.0001 & 0.9716 $\pm$ 0.0002          & 0.9589 $\pm$ 0.0002          & 0.9515 $\pm$ 0.0003          & 0.9689 $\pm$ 0.0005          & 0.9620 $\pm$ 0.0001          & \textbf{0.9734 $\pm$ 0.0002} \\ \hline
			\textbf{Ecoli1}                                     & 0.7936 $\pm$ 0.0071                            & 0.8452 $\pm$ 0.0044 & 0.8721 $\pm$ 0.0057          & \textbf{0.8725 $\pm$ 0.0040} & 0.8654 $\pm$ 0.0028          & 0.8699 $\pm$ 0.0030          & 0.8657 $\pm$ 0.0030          & \textbf{0.8588 $\pm$ 0.0061} \\ \hline
			\textbf{Yeast3}                                     & \textbf{0.7292 $\pm$ 0.0037}                   & 0.8108 $\pm$ 0.0021 & 0.8015 $\pm$ 0.0010          & 0.8142 $\pm$ 0.0009          & 0.7934 $\pm$ 0.0011          & \textbf{0.8221 $\pm$ 0.0004} & 0.7949 $\pm$ 0.0014          & \textbf{0.7986 $\pm$ 0.0003} \\ \hline
			\textbf{Pageblock}                                  & 0.4740 $\pm$ 0.0651                            & 0.9228 $\pm$ 0.0046 & 0.9345 $\pm$ 0.0049          & \textbf{0.9753 $\pm$ 0.0012} & 0.9460 $\pm$ 0.0033          & 0.9345 $\pm$ 0.0049          & 0.9729 $\pm$ 0.0002          & \textbf{0.9428 $\pm$ 0.0130} \\ \hline
			\textbf{Glass5}                                     & 0.6111 $\pm$ 0.1382                            & 0.6111 $\pm$ 0.1382 & 0.6111 $\pm$ 0.1382          & 0.6111 $\pm$ 0.1382          & 0.6111 $\pm$ 0.1382          & 0.6020 $\pm$ 0.1406          & 0.6020 $\pm$ 0.1406          & \textbf{0.6222 $\pm$ 0.1362} \\ \hline
			\textbf{Yeast5}                                     & 0.5080 $\pm$ 0.0183                            & 0.7682 $\pm$ 0.0019 & 0.7615 $\pm$ 0.0040          & 0.7804 $\pm$ 0.0015          & 0.6342 $\pm$ 0.0041          & 0.7613 $\pm$ 0.0034          & 0.7943 $\pm$ 0.0010          & \textbf{0.8004 $\pm$ 0.0027} \\ \hline
			\textbf{Yeast6}                                     & 0.2272 $\pm$ 0.0439                            & 0.5544 $\pm$ 0.0066 & 0.5630 $\pm$ 0.0092          & 0.6059 $\pm$ 0.0147          & 0.5388 $\pm$ 0.0123          & 0.6069 $\pm$ 0.0168          & 0.5002 $\pm$ 0.0091          & \textbf{0.6350 $\pm$ 0.0049} \\ \hline
			\textbf{Abalone}                                    & 0.0 $\pm$ 0.0                                  & 0.1201 $\pm$ 0.0001 & 0.1238 $\pm$ 0.0001          & 0.0830 $\pm$ 0.0017          & 0.1106 $\pm$ 0.0038          & 0.1020 $\pm$ 0.0027          & 0.1250 $\pm$ 0.0001          & \textbf{0.1465 $\pm$ 0.0011} \\ \hline
			\textbf{Anger-Anxiety}                              & 0.9218 $\pm$ 0.0007                            & 0.9307 $\pm$ 0.0002 & 0.9417 $\pm$ 0.0002          & 0.9417 $\pm$ 0.0002          & 0.9220 $\pm$ 0.0011          & \textbf{0.9532 $\pm$ 0.0006} & 0.9391 $\pm$ 0.0003          & \textbf{0.9532 $\pm$ 0.0004} \\ \hline
			\textbf{Anger-Disgust}                              & 0.8479 $\pm$ 0.0063                            & 0.8669 $\pm$ 0.0064 & 0.8631 $\pm$ 0.0063          & 0.8631 $\pm$ 0.0063          & 0.8771 $\pm$ 0.0039          & \textbf{0.8972 $\pm$ 0.0092} & 0.8594 $\pm$ 0.0067          & 0.8669 $\pm$ 0.0064          \\ \hline
			\textbf{Anger-Happy}                                & 0.6234 $\pm$ 0.0084                            & 0.6813 $\pm$ 0.0052 & 0.6867 $\pm$ 0.0054          & 0.6845 $\pm$ 0.0072          & 0.6724 $\pm$ 0.0101          & 0.7021 $\pm$ 0.0027          & 0.6989 $\pm$ 0.0074          & \textbf{0.7071 $\pm$ 0.0073} \\ \hline
			\textbf{Anger-Sad}                                  & 0.9804 $\pm$ 0.0007                            & 0.9724 $\pm$ 0.0030 & \textbf{0.9864 $\pm$ 0.0007} & \textbf{0.9864 $\pm$ 0.0007} & 0.9599 $\pm$ 0.0029          & \textbf{0.9907 $\pm$ 0.0001} & \textbf{0.9864 $\pm$ 0.0007} & \textbf{0.9864 $\pm$ 0.0007} \\ \hline
			\textbf{Anxiety-Disgust}                            & 0.7514 $\pm$ 0.0251                            & 0.7989 $\pm$ 0.0094 & 0.8096 $\pm$ 0.0100          & 0.8096 $\pm$ 0.0100          & 0.8096 $\pm$ 0.0100          & \textbf{0.8220 $\pm$ 0.0124} & 0.8096 $\pm$ 0.0100          & \textbf{0.8124 $\pm$ 0.0091} \\ \hline
			\textbf{Boredom-Disgust}                            & 0.8385 $\pm$ 0.0068                            & 0.8188 $\pm$ 0.0035 & 0.8584 $\pm$ 0.0034          & \textbf{0.8774 $\pm$ 0.0032} & \textbf{0.8906 $\pm$ 0.0028} & 0.8501 $\pm$ 0.0014          & 0.8737 $\pm$ 0.0037          & \textbf{0.8774 $\pm$ 0.0032} \\ \hline
			\textbf{Happy-Disgust}                              & 0.8806 $\pm$ 0.0078                            & 0.8787 $\pm$ 0.0100 & \textbf{0.9027 $\pm$ 0.0050} & 0.8850 $\pm$ 0.0087          & 0.9157 $\pm$ 0.0039          & \textbf{0.9022 $\pm$ 0.0042} & 0.8998 $\pm$ 0.0096          & \textbf{0.9027 $\pm$ 0.0050} \\ \hline
			\textbf{Neutral-Anger}                              & 0.9287 $\pm$ 0.0037                            & 0.9180 $\pm$ 0.0048 & 0.9313 $\pm$ 0.0040          & 0.9185 $\pm$ 0.0026          & 0.9125 $\pm$ 0.0083          & \textbf{0.9392 $\pm$ 0.0021} & 0.9287 $\pm$ 0.0037          & 0.9287 $\pm$ 0.0037          \\ \hline
			\textbf{Neutral-Disgust}                            & 0.8072 $\pm$ 0.0058                            & 0.8231 $\pm$ 0.0030 & \textbf{0.8584 $\pm$ 0.0040} & 0.8384 $\pm$ 0.0074          & \textbf{0.8584 $\pm$ 0.0040} & 0.8547 $\pm$ 0.0044          & \textbf{0.8584 $\pm$ 0.0040} & 0.8405 $\pm$ 0.0010          \\ \hline
			\textbf{Sad-Disgust}                                & 0.9264 $\pm$ 0.0048                            & 0.9264 $\pm$ 0.0048 & 0.9264 $\pm$ 0.0048          & 0.9264 $\pm$ 0.0048          & 0.9264 $\pm$ 0.0048          & 0.9264 $\pm$ 0.0048          & 0.9264 $\pm$ 0.0048          & \textbf{0.9446 $\pm$ 0.0055} \\ \hline
			\textbf{AD-MCI}                                     & 0.5190 $\pm$ 0.0133                            & 0.5300 $\pm$ 0.0067 & 0.5401 $\pm$ 0.0052          & 0.5345 $\pm$ 0.0043          & 0.5467 $\pm$ 0.0061          & 0.5404 $\pm$ 0.0088          & 0.5440 $\pm$ 0.0096          & \textbf{0.5563 $\pm$ 0.0087} \\ \hline
			\textbf{HC-MCI}                                     & 0.5686 $\pm$ 0.0027                            & 0.5846 $\pm$ 0.0018 & 0.5767 $\pm$ 0.0040          & 0.5770 $\pm$ 0.0010          & 0.5940 $\pm$ 0.0033          & 0.5919 $\pm$ 0.0013          & 0.5840 $\pm$ 0.0023          & \textbf{0.6003 $\pm$ 0.0053} \\ \hline
			\textbf{04clover5z-600-5-70-BI}                     & 0.0 $\pm$ 0.0                                  & 0.6563 $\pm$ 0.0031 & 0.6680 $\pm$ 0.0013          & 0.6712 $\pm$ 0.0036          & 0.6665 $\pm$ 0.0027          & 0.6006 $\pm$ 0.0161          & \textbf{0.6856 $\pm$ 0.0024} & \textbf{0.6577 $\pm$ 0.0080} \\ \hline
			\textbf{04clover5z-600-5-60-BI}                     & 0.0 $\pm$ 0.0                                  & 0.6872 $\pm$ 0.0057 & 0.6966 $\pm$ 0.0007          & 0.6909 $\pm$ 0.0017          & 0.7051 $\pm$ 0.0046          & 0.6304 $\pm$ 0.0054          & \textbf{0.7157 $\pm$ 0.0035} & \textbf{0.7141 $\pm$ 0.0023} \\ \hline
			\textbf{04clover5z-600-5-50-BI}                     & 0.0 $\pm$ 0.0                                  & 0.6913 $\pm$ 0.0059 & 0.7281 $\pm$ 0.0036          & 0.7191 $\pm$ 0.0018          & 0.7054 $\pm$ 0.0029          & 0.6392 $\pm$ 0.0068          & 0.7156 $\pm$ 0.0015          & \textbf{0.7285 $\pm$ 0.0033} \\ \hline
			\textbf{04clover5z-600-5-30-BI}                     & 0.0 $\pm$ 0.0                                  & 0.6887 $\pm$ 0.0028 & 0.6895 $\pm$ 0.0038          & 0.6972 $\pm$ 0.0023          & 0.7083 $\pm$ 0.0009          & 0.6871 $\pm$ 0.0021          & 0.7083 $\pm$ 0.0023          & \textbf{0.7153 $\pm$ 0.0005} \\ \hline
			\textbf{04clover5z-600-5-0-BI}                      & 0.0 $\pm$ 0.0                                  & 0.6961 $\pm$ 0.0009 & 0.7038 $\pm$ 0.0021          & 0.7331 $\pm$ 0.0017          & 0.7370 $\pm$ 0.0017          & 0.7201 $\pm$ 0.0011          & 0.7375 $\pm$ 0.0012          & \textbf{0.7478 $\pm$ 0.0007} \\ \hline
		\end{tabular}
	}
	\label{table:F2}
\end{table*}

\subsection{Experimental Setup}
\def\C{$\lambda$}
We compare the performance of our proposed approach with various state-of-the-art oversampling techniques, namely, Random Oversampling (ROS), SMOTE \cite{SMOTE}, BSMOTE-1 \cite{BSMOTE}, BSMOTE-2 \cite{BSMOTE}, SVMSMOTE \cite{SVMSMOTE}, ADASYN \cite{ADASYN}, and baseline (i.e. no oversampling). In order to implement the oversampling methods, we use the widely popular {\em imbalanced-learn} toolbox. For each of the algorithms, we employ a grid-search to tune the neighborhood and other unique parameters to optimize the classification performance for each of the comparative algorithms. For the proposed algorithm, we adopt a similar mechanism to tune the number of clusters ($C$) and the neighborhood parameter ($K$) to achieve optimal performance. We use the Support Vector Machine (SVM) classifier with the radial basis function (RBF) kernel to conduct the experiments. In order to have a fair comparison, the classifier parameters are kept the same for each of the oversampling techniques. Specifically, we set the regularization parameter '\C' of the SVM classifier to a value of $1$ and all other unique parameters are set to the default values as specified in the {\em scikit-learn} machine learning library \cite{scikit-learn}. The features are normalized by subtracting the mean and then scaling to unit variance (i.e zero mean and unit variance) before training the SVM classifier. For comparing the performance, we adopt the F-measure metric which is defined as,

\[
F_\beta = (1 + \beta^2) \times \frac{Precision \times Recall}{(\beta^2)\times Precision + Recall}
\]

$Precision$ is defined as a ratio of the number of minority class samples correctly classified to the total number of samples classified as minority class, while $Recall$ is the ratio of the number of minority class samples that are correctly classified to the total number of minority class samples in the test set. We use the $F_1$ $(\beta = 1)$ and $F_2$ $(\beta = 2)$ scores to compare the different algorithms. For all the experiments, we perform a $5$-fold cross-validation by ensuring that the imbalance ratio across each fold is maintained the same. All results are reported as mean $\pm$ variance.

\begin{table*}[h]
	\caption{Comparison of Minority Class and Overall Accuracy Scores obtained by using the proposed approach and the baseline (Improvement, detriment and no change with respect to baseline are denoted by $\uparrow$, $\downarrow$ and $\updownarrow$ respectively)}
	\centering
	\scalebox{1.0}{
		\begin{tabular}{|l|c|c|c|c|c|c|}
			\hline
			& \multicolumn{3}{c|}{\textbf{Minority Accuracy}} & \multicolumn{3}{c|}{\textbf{Overall Accuracy}} \\ \cline{2-7} 
			\multirow{-2}{*}{{\color[HTML]{000000} \textbf{Classification Task}}}                                          & \textbf{Baseline}   & \textbf{Proposed}   & \textbf{Difference}   & \textbf{Baseline}   & \textbf{Proposed}  & \textbf{Difference}  \\\hline  \hline
			\textbf{Pima}                                       & 0.5410     & 0.7015     & 0.1605 $\uparrow$       & 0.7669     & 0.7630     & -0.0039 $\downarrow$     \\ \hline
			\textbf{Glass0}                                     & 0.6142     & 0.8857     & 0.2715 $\uparrow$      & 0.7897     & 0.8084     & 0.0187 $\uparrow$     \\ \hline
			\textbf{Vehicle0}                                   & 0.9396     & 0.9899     & 0.0503 $\uparrow$      & 0.9680     & 0.9751     & 0.0071 $\uparrow$    \\ \hline
			\textbf{Ecoli1}                                     & 0.7792     & 0.8442     & 0.0650 $\uparrow$       & 0.9196     & 0.9077     & -0.0119 $\downarrow$    \\ \hline
			\textbf{Yeast3}                                     & 0.7116     & 0.8282     & 0.1166 $\uparrow$       & 0.9508     & 0.9420     & -0.0088 $\downarrow$    \\ \hline
			\textbf{Pageblock}                                  & 0.4285     & 0.9285     & 0.5000 $\uparrow$       & 0.9661     & 0.9957     & 0.0296 $\uparrow$      \\ \hline
			\textbf{Glass5}                                     & 0.5555     & 0.5555     & 0.0000  $\updownarrow$     & 0.9766     & 0.9813     & 0.0047 $\uparrow$ \\ \hline
			\textbf{Yeast5}                                     & 0.4772     & 0.8864     & 0.4092 $\uparrow$       & 0.9804     & 0.9750     & -0.0054 $\downarrow$     \\ \hline
			\textbf{Yeast6}                                     & 0.2000     & 0.7428     & 0.5428 $\uparrow$       & 0.9804     & 0.9676     & -0.0128 $\downarrow$     \\ \hline
			\textbf{Abalone}                                    & 0.0000     & 0.5938     & 0.5938 $\uparrow$       & 0.9930     & 0.8732     & -0.1198 $\downarrow$     \\ \hline
			\textbf{Anger-Anxiety}                              & 0.9420     & 0.9565     & 0.0145 $\uparrow$       & 0.9218     & 0.9532     & 0.0314  $\uparrow$     \\ \hline
			\textbf{Anger-Disgust}                              & 0.8260     & 0.8478     & 0.0218 $\uparrow$       & 0.9421     & 0.9481     & 0.0060 $\uparrow$     \\ \hline
			\textbf{Anger-Happy}                                & 0.5915     & 0.6901     & 0.0986 $\uparrow$       & 0.8026     & 0.8233     & 0.0207 $\uparrow$      \\ \hline
			\textbf{Anger-Sad}                                  & 0.9838     & 0.9838     & 0.0000 $\updownarrow$      & 0.9842     & 0.9947     & 0.0105 $\uparrow$      \\ \hline
			\textbf{Anxiety-Disgust}                            & 0.7173     & 0.7826     & 0.0653 $\uparrow$      & 0.8608     & 0.8956     & 0.0348 $\uparrow$     \\ \hline
			\textbf{Boredom-Disgust}                            & 0.8260     & 0.8696     & 0.0436 $\uparrow$      & 0.9132     & 0.9292     & 0.0160 $\uparrow$      \\ \hline
			\textbf{Happy-Disgust}                              & 0.8695     & 0.8913     & 0.0218 $\uparrow$       & 0.9235     & 0.9405     & 0.0170 $\uparrow$     \\ \hline
			\textbf{Neutral-Anger}                              & 0.9240     & 0.9240     & 0.0000 $\updownarrow$       & 0.9516     & 0.9516     & 0.0000 $\updownarrow$     \\ \hline
			\textbf{Neutral-Disgust}                            & 0.7826     & 0.8261     & 0.0435 $\uparrow$       & 0.8960     & 0.9040     & 0.0080 $\uparrow$      \\ \hline
			\textbf{Sad-Disgust}                                & 0.9130     & 0.9347     & 0.0217 $\uparrow$       & 0.9627     & 0.9718     & 0.0091 $\uparrow$     \\ \hline
			\textbf{AD-MCI}                                     & 0.5000     & 0.5431     & 0.0431 $\uparrow$      & 0.7493     & 0.7521     & 0.0028 $\uparrow$      \\ \hline
			\textbf{HC-MCI}                                     & 0.5517     & 0.5862     & 0.0345 $\uparrow$       & 0.7569     & 0.7710     & 0.0141 $\uparrow$    \\ \hline
			\textbf{04clover5z-600-5-70-BI}                     & 0.0000     & 0.7600     & 0.7600 $\uparrow$       & 0.8333     & 0.7583     & -0.0750 $\downarrow$    \\ \hline
			\textbf{04clover5z-600-5-60-BI}                     & 0.0000     & 0.7080     & 0.7080 $\uparrow$       & 0.8333     & 0.7433     & -0.0900 $\downarrow$    \\ \hline
			\textbf{04clover5z-600-5-50-BI}                     & 0.0000     & 0.7180     & 0.7180 $\uparrow$       & 0.8333     & 0.7500     & -0.0833 $\downarrow$    \\ \hline
			\textbf{04clover5z-600-5-30-BI}                     & 0.0000     & 0.7100     & 0.7100 $\uparrow$       & 0.8333     & 0.7416     & -0.0917 $\downarrow$    \\ \hline
			\textbf{04clover5z-600-5-0-BI}                      & 0.0000     & 0.7200     & 0.7200 $\uparrow$       & 0.8333     & 0.7650     & -0.0683 $\downarrow$    \\ \hline
		\end{tabular}
	}
	\label{table:accuracy}
\end{table*}

\subsection{Parameter Selection}

In the proposed algorithm, we have a set of $5$ parameters that are to be tuned for optimal performance. For our experiments, the values chosen for these parameters are as follows:

\begin{itemize}
	\item The neighborhood parameter ($K$) is varied between $2$ and $10$.
	
	\item The number of clusters ($C$) is varied based on the number of majority ($M$) and minority ($N$) samples as $(M-N)\times\eta$, where $\eta \in \{0.1, 0.2, 0.3, 0.4, 0.5, 0.6, 0.7, 0.8, 0.9, 1\}$.
	
	\item The number of samples to be picked up between two adjacent samples ($r$) is fixed at $10$.
	
	\item The probability threshold ($p_t$) is fixed at $0.5$.
	
	\item The weight threshold ($w_t$) is fixed at $0.5$.
\end{itemize}

\subsection{Results and Analysis}

In Tables \ref{table:F1} and \ref{table:F2}, the $F_1$ and $F_2$ scores of various algorithms with the SVM classifier on 27 different binary classification tasks are presented with the the best result in each task highlighted in boldface. It can be inferred from the results that, upon comparison with the baseline algorithm (i.e. no sampling), for most of the datasets the proposed approach significantly boosts the classification results, clearly evident by higher $F_1$ and $F_2$ scores. Even on datasets for which the baseline algorithm itself provides reasonable classification scores (such as Vehicle0, Anger-Sad, Anger-Anxiety), the proposed algorithm provides a slight improvement in the classifier performance. In addition to improved performance with respect to the baseline, the proposed adaptive oversampling technique outperforms other state-of-the-art oversampling techniques by a wide margin in most of the datasets. In case of highly imbalanced datasets, the proposed technique offers a huge improvement in performance, compared to the other methods. It is to be noted that our proposed algorithm outperforms the other techniques in $22$ out of $27$ times in terms of the $F_1$ score metric and in $16$ out of $27$ times in terms of the $F_2$ score metric. Further, the minority class and overall accuracies obtained by using the proposed algorithm and the baseline is presented in Table \ref{table:accuracy}. From the experimental
results it can be concluded that the algorithm not only boosts the minority classification performance but also performs reasonably well in overall classification
thereby verifying the usability of the oversampling technique proposed in this paper. The proposed technique can also be applied to multi-class imbalanced scenarios by modifying the algorithm suitably or by the use of appropriate boosting methods \cite{Tanha2020BoostingMF}.

\section{Conclusion} \label{sec:conclusion}

Classification in imbalanced datasets is a challenging problem and an active research area in the field of machine learning. In this paper, we propose a novel oversampling technique in order to mitigate the bias of conventional classifiers towards the majority class in such scenarios. We identify that the Synthetic Minority Oversampling Technique (SMOTE), has the drawback that the important local distribution characteristics of the data are ignored. 
To deal with this, we propose a three step sampling technique that adaptively selects specific data points during the process of oversampling. For this purpose, first a specified number of synthetic minority samples (much higher than required to achieve a balanced distribution) are generated by using the conventional SMOTE. Then, the generated samples are clustered by using GMM-based clustering algorithm. Each of the clusters are assigned a weight based on the majority class distribution which is then used to choose the final set of synthetic samples from each of the clusters. It is to be noted that the majority class distribution is used in assigning a weight to each of the clusters as it allows the algorithm to employ the more diverse distribution compared to the minority class distribution. The resulting training class data, post the application of the proposed oversampling technique, 
not only produces a balanced representation of the data distribution but also ensures that the synthetically 
generated samples are adequately diverse and representative of the original distribution. 
We validate the efficacy of our approach by using the $F_1$ and $F_2$ score as performance metrics on $27$ binary classification tasks with varying degrees of imbalance, attributes and domains.  

\bibliographystyle{IEEEtran}
\bibliography{ICPR2020}

\end{document}